\documentclass[runningheads]{llncs}
\usepackage[T1]{fontenc}
\usepackage{graphicx}
\usepackage{xcolor}
\usepackage{amsmath}
\usepackage{amssymb}
\usepackage{tabularx}
\usepackage{hyperref}
\usepackage{listings}
\usepackage{enumerate}
\usepackage{multirow}

\usepackage{color}

\usepackage[normalem]{ulem}
\newcommand{\old}[1]{\textcolor{red}{\ifmmode\text{\sout{\ensuremath{#1}}}\else\sout{#1}\fi}}

\begin{document}

\title{Bridging between LegalRuleML and TPTP for Automated Normative Reasoning\\(extended version)\thanks{The second author acknowledges financial support from the Luxembourg National Research Fund (FNR) under grant CORE AuReLeE (C20/IS/14616644).}}
\titlerunning{Bridging between LegalRuleML and TPTP}
%
\author{Alexander Steen\inst{1}\orcidID{0000-0001-8781-9462} \\ \and
David Fuenmayor\inst{2,3}\orcidID{0000-0002-0042-4538}}
\authorrunning{A. Steen and D. Fuenmayor}
%
\institute{University of Greifswald, Germany \\
\email{alexander.steen@uni-greifswald.de} \and
University of Luxembourg, Luxembourg \\
\email{david.fuenmayor@uni.lu} \and
University of Bamberg, Germany 
}
\maketitle              
%

\begin{abstract}
LegalRuleML is a comprehensive XML-based representation framework for modeling and exchanging normative rules.
The TPTP input and output formats, on the other hand, are general-purpose standards for the interaction with
automated reasoning systems. In this paper we provide a bridge between the two communities by
(i) defining a logic-pluralistic normative reasoning language based on the TPTP format,
(ii) providing a translation scheme between relevant fragments of LegalRuleML and this language,
and (iii) proposing
a flexible architecture for automated normative reasoning based on this translation.
We exemplarily instantiate and demonstrate the approach with three different normative logics.

\end{abstract}


\section{Introduction}

Automated theorem proving (ATP) systems are computer programs that, given a set $A$ of assumptions and a 
conjecture $C$ as input, try to prove that $C$ is a logical consequence of $A$,
i.e., that it is impossible for $C$ to be false whenever every formula from $A$ holds.
ATP systems conduct the whole reasoning process automatically, so that no user interaction
is necessary during proof search. This is in contrast to interactive theorem provers, also called proof assistants, which provide more sophisticated user interfaces in order to take advantage of human ingenuity during an interactive process of proof (re-)construction.

In normative reasoning, logical formalisms are employed to represent
and reason about different notions of norms, including obligations,
permissions and prohibitions. In automated normative reasoning, the goal
is hence to automate the reasoning process in the context of normative discourse by employing suitable
logical systems.
LegalRuleML~\cite{AthanBGPPW13,AthanGPPW15} is a comprehensive XML-based representation framework
for modeling and exchanging normative rules, e.g., legal norms originating
from national laws of some particular country.
The LegalRuleML standard comes with fine-grained and expressive means for
representing (possibly legal) norms in an \textit{isomorphic} \cite{bench1992isomorphism} fashion with respect to their original source(s).
At the same time, 
LegalRuleML is \textit{semantically underspecified}, in the sense that it deliberately does not prescribe a specific logic (or semantics) in which the represented norms are to be interpreted.

LegalRuleML has been employed by Robaldo et al.\ to provide an exhaustive formalization
of the General Data Protection Regulation (GDPR)~\cite{DBLP:journals/jolli/RobaldoBPRML20}.
Palmirani and Governatori combine LegalRuleML with further technologies and approaches to present an integrated
framework for compliance checking with legal rules, but also focusing on GDPR applications~\cite{DBLP:conf/jurix/PalmiraniG18}.
Still, there exist only comparably few
systems that, in fact, automate reasoning processes based on normative
knowledge.
Notable examples are provided by Liu et al.\ who interpret legal norms
in a defeasible deontic logic and provide automation for it~\cite{LiuIG21},
and the SPINdle prover~\cite{DBLP:conf/ruleml/LamG09} for propositional (modal) defeasible reasoning that has been used in multiple works 
in the normative application domain.

In contrast, there are many general-purpose ATP systems available for classical logics, e.g.,
for propositional logic, first-order predicate logic, and more recently for higher-order logic.
These systems are being continuously improved and are increasingly becoming more effective,
as witnessed by the results of the annual ATP system competition CASC~\cite{Sut16}.
The development of general-purpose ATP systems for normative reasoning is, on the other hand,
complicated by the fact that there is no single logic acting as the
de-facto standard formalism for normative reasoning.
In general, the design and implementation of practically effective ATP systems is a non-trivial task and very laborious; and so it is easy to see that developing custom ATP systems for each distinct normative formalism is a quite unfeasible undertaking, in particular since, as witnessed in deontic logic (see Sect.~\ref{subsec:deontic-logics}), those formalisms behave as moving targets.

In this paper, we therefore propose to employ for this task general-purpose ATP
systems for classical higher-order logic, and thus to reduce normative reasoning tasks to classical ATP problems in a quite general way.
For this, we bridge between LegalRuleML and the TPTP language standard for ATP systems~\cite{Sut17}, so that
any TPTP-compliant ATP system for higher-order logic can be reused as a reasoning backend for a wide range
of normative logics.

The contributions of this paper are as follows:
\begin{itemize}
    \item We define a logic-pluralistic domain specific language (DSL)
          for normative reasoning with TPTP-compliant ATP systems.
    \item We show how the DSL can be mechanically translated into
          ATP reasoning problems in different concrete normative logics.
    \item We describe a reasoning architecture that provides flexible means
          of automation for these normative logics.
    \item We present a prototypical implementation of the whole
          reasoning tool chain that is available as open-source
          code.
\end{itemize}

The remainder of this paper is structured as follows:
In Sect.~\ref{sec:prelim} we briefly survey the TPTP and the LegalRuleML standards, together with a very brief exposition
of deontic logics as specific systems for normative reasoning.
In Sect.~\ref{sec:logic-pluralism}, we discuss the role of logical pluralism in normative reasoning, which is
one of the key motivations of this work.
Subsequently, Sect.~\ref{sec:normativeKR} presents the utilization of a TPTP format for normative rules.
In Sect.~\ref{sec:backends} we then present a flexible and uniform approach for automating normative reasoning using general purpose ATP systems.
Finally, Sect.~\ref{sec:conclusion} concludes and sketches further work.

\paragraph{Related work.}
Automation approaches by translation have also been studied by Lam and Hashmi, where they translate LegalRuleML
statements into a defeasible modal logic for which an automated reasoning tool exists~\cite{LamH19}. Their approach
is, however, fixed to one specific logic formalism as opposed to the logic-pluralistic view that we put forward
in this work. Similarly, Boley et al.\ translate RuleML information to modal logic in TPTP format~\cite{BoleyBLS16}. However,
simple modal logics are not fully adequate for normative reasoning, see the brief discussion in Sect.~\ref{subsec:deontic-logics}.
Our approach is in line with the LogiKEy methodology proposed by Benzmüller et al.~\cite{BenzmullerPT20},
which makes use of expressive higher-order logics for flexibly encoding, reasoning, and experimenting with normative theories.

\section{Preliminaries \label{sec:prelim}}

\subsection{The TPTP Infrastructure for ATP systems}\label{subsec:tptp}
The \emph{Thousands of Problems for Theorem Proving} (TPTP) library and infrastructure~\cite{Sut17} is the core
platform for contemporary ATP system development and evaluation. It provides (i) a comprehensive collection
of benchmark problems for ATP systems; (ii) a set of utility tools for problem and solution inspection, pre- and post-processing,
and verification; and (iii) a comprehensive syntax standard for ATP system input and output. 
Currently, the TPTP contains approx.\ 25.000 reasoning problems from more than 50 different application domains. 

The TPTP specifies different ATP system languages varying in their expressivity: The \emph{first-order form} (FOF)
represents unsorted first-order logic, the \emph{typed first-order form} (TFF) represents many-sorted first-order logic~\cite{SutcliffeSCB12},
and \emph{typed higher-order form} (THF) represents classical higher-order logic~\cite{SutcliffeB10}. All input formats are in ASCII plain text,
human-readable and follow Prolog language conventions for simple parsing. An ATP problem generally consists of
symbol declarations (if the language is typed), contextual definitions and premises of the reasoning
task (usually referred to as \emph{axioms}), and a conjecture that is to be proved
or refuted in the given context. The core building block of the ATP problem files
in TPTP languages are
so-called \emph{annotated formulas} of form \ldots \smallskip

\mbox{} \quad {\em language}{\tt (}{\em name}{\tt ,}{\em role}{\tt ,}{\em formula}[{\tt ,}{\em source}[{\tt ,}{\em annotations}]]{\tt ).}
\smallskip

Here, {\em language} is a three-letter identifier for the intended language in which the annotated formula is expressed
({\tt fof}, {\tt tff} or {\tt thf}). The {\em name} is a unique identifier for referencing to the annotated formula
but has no other effect on the interpretation of it. The {\em role} field specifies whether the {\em formula} should be interpreted
as an assumption (role {\tt axiom}), a type declaration (role {\tt type}), a definition (role {\tt definition}) or as
formula to be proved (role {\tt conjecture}).\footnote{In Sect.~\ref{sec:normativeKR} we will discuss a further role, {\tt logic}, used for \textit{logic specifications.}} The {\em formula} is an ASCII representation of the respective logical
expression, where predicate and function symbols are denoted by strings that begin with a lower-case letter,
variables are denoted by strings starting with an upper-case letter, the logical connectives 
$\neg$, $\land$, $\lor$, $\rightarrow$, $\leftrightarrow$ are represented by {\tt {\char`\~}}, {\tt \&}, {\tt |}, {\tt =>} and {\tt <=>}, respectively.
Quantifiers $\forall$ and $\exists$ are expressed by {\tt !} and {\tt ?}, respectively, followed by a list of
variables bound by it. The TPTP defines several interpreted constants starting with a {\tt \$}-sign, 
including {\tt \$true} and {\tt \$false} for truth and falsehood, respectively. In typed languages, such as
TFF and THF, the type {\tt \$i} represents the type of individuals and {\tt \$o} is the type of Booleans.
In TFF, explicit types of symbols may be dropped and default to n-ary function types {\tt (\$i * \ldots\ * \$i) > \$i} and n-ary predicate types {\tt (\$i * \ldots\ * \$i) > \$o} depending on their occurrence. In TFF, arguments are applied using parentheses as in
\verb|f(a,b)|, where \verb|f| is some function symbol, and \verb|a| and \verb|b| are some terms (of appropriate type).
In THF, however, arguments are applied in a curried fashion with an explicit application operator \verb|@|, as in
\verb|f @ a @ b|.
Finally, the {\em source} and {\em annotations} are optional and uninterpreted extra-logical information that can be assigned to the
annotated formula, e.g., about its origin, its relevance, or other properties.
An example in TFF is as follows:
\begin{verbatim}
  tff(union_def, axiom, ! [S, T, X]: (
                            member(X, union(S,T)) <=>
                           ( member(X, S) | member(X, T) ) ),
                        source('definitions.ax'),
                        [relevance(1.0)]).
\end{verbatim}
In this example, a TFF annotated formula of name {\tt union\_def} is given that describes an axiom 
giving a fundamental property of set  union and some auxiliary information about it.
A complete description of the TPTP input languages, including the syntax BNF,
is provided at the TPTP web page\footnote{\url{http://tptp.org}}. For further information about available tools,
problems sets and an extensive description of the different  languages, we refer to the
literature~\cite{Sut17}.

\subsection{Deontic Logics and LegalRuleML} \label{subsec:deontic-logics}
Deontic logics are logical systems intended to formally represent normative notions, such as obligations, permissions and prohibitions, their relationships, and their properties~\cite{handbookDeontic}. An early deontic logic, today still referred to as standard deontic logic (SDL), is based on simple modal logic \textbf{D}. In this context,
the modal operators are usually denoted $O$ (for obligation) and $P$ (for permission), where $O\varphi \leftrightarrow \neg P \neg \varphi$ holds, and
every instance of $O\varphi \rightarrow P\varphi$ is validated~\cite{hilpinen2013deontic}.

Note that the term \emph{standard deontic logic} is a misnomer because the
system has many undesired properties, as it is vulnerable to many well-known deontic logic `paradoxes'.
In normative reasoning contexts, usually other deontic logics are employed today. There exist several different kinds of them; we mention just a few: Dyadic deontic logics specifically
address conditional norms of the form $O(\varphi|\psi)$
(read: \emph{It ought to be $\varphi$ given $\psi$})~\cite{prakken1997dyadic},
defeasible deontic logics address non-monotonic reasoning patterns with
defeasible norms~\cite{governatori2012possible}, and norm-based deontic logics model
norms separately from factual (non-deontic) logical formulas~\cite{makinson2000input}.

A prominent example illustrating the shortcomings of SDL related to conditional norms is Chisholm's paradox~\cite{chisholm1963}, paraphrased as follows:

{
\noindent {\itshape Assume that your neighbors are in trouble (and you like them), then \dots}
\begin{enumerate}[(1)]
\item {\itshape You ought to go help your neighbors.}
\item {\itshape If you go help your neighbors, you ought to tell them you are coming over.}
\item {\itshape If you do not go help your neighbors, you ought not to tell them you are coming over.}
\item {\itshape You do not go help your neighbors.}
\end{enumerate}
}

The sentences (1) -- (4) above appear to describe a plausible situation, and, intuitively, they also constitute a both logically consistent and independent set of sentences. Hence, arguably, an adequate formalization should respect these constraints.
Chisholm's paradox here mainly serves as a running example that highlights the significant effects of interpreting
normative information under different logical systems. 

\begin{table}
\centering
\noindent\begin{tabular}{p{50mm}|c|c|c|c}
\multicolumn{1}{c|}{\textbf{Natural language}} & \textbf{SDL-v1} & \textbf{SDL-v2} & \textbf{SDL-v3} & \textbf{DDL} \\ \hline
You ought to go help \newline your neighbors & $O\, h$ & $O\, h$ & $O\, h$ & $O\, h$ \\ \hline
If you go help your neighbors, \newline you ought
to tell them you are \newline coming over & $O\,(h \rightarrow  t)$ & $O\,(h \rightarrow  t)$ & $h \rightarrow  O\,t$ &  $O\,(t|h)$ \\ \hline
If you do not go help your \newline neighbors, you ought not to tell \newline them you are coming over & $\neg h \rightarrow  O\,\neg t$ & $O\,(\neg h \rightarrow \neg t)$ & $\neg h \rightarrow  O\,\neg t$ & $O\,(\neg t|\neg h)$ \\ \hline
You do not go help your neighbors & $\neg h$ & $\neg h$ & $\neg h$ & $\neg h$ \\ 
\end{tabular}
\smallskip

\caption{Some possible formalizations of Chisholm's paradox.}
\label{table:chisholm-interpretations}
\end{table}
Table \ref{table:chisholm-interpretations} shows several different interpretations for Chisholm's scenario; three of them formalized using SDL, and the fourth formalized using a dyadic deontic logic (DDL)
where $h$ represents ``helping your neighbors'' and $t$ represents ``telling them you are coming over''. As it happens, the set of formulas corresponding to the first SDL-formalization variant (SDL-v1) is inconsistent, thus allowing the derivation
of every formula and, in particular, every obligation (e.g., $O\, k$ where $k$ could represent ''killing your neighbor'').
In fact, the next two SDL-formalizations are not logically  independent, and thus inadequate, see~\cite[\S 8.5]{hilpinen2013deontic} for a discussion.
In dyadic deontic logics the conditional norms from above are represented using dyadic obligation
operators as in $O(t|h)$ resp. $O(\neg t|\neg h)$ instead of material implications. These logic systems (e.g.\ {\AA}qvist's system \textbf{E}~\cite{aaqvist1984deontic,DBLP:journals/rsl/Parent15} and Carmo \& Jones' system~\cite{carmo2002deontic}) are specifically conceived in order to remedy shortcomings of SDL in addressing the so-called `paradoxes' of deontic logic related to conditional obligations. Unsurprisingly, no logic formalism has yet been found which successfully addresses all of the many different deontic paradoxes and deficiencies, see~\cite[\S 8]{hilpinen2013deontic} for an exhaustive
overview.

\paragraph{LegalRuleML.}
LegalRuleML~\cite{AthanBGPPW13} is a comprehensive XML-based representation framework for modeling and exchanging normative rules.
It extends the general RuleML standard~\cite{DBLP:conf/ruleml/BoleyPS10} with specialized concepts and features for normative rules, legal contexts, interpretations, etc.

In LegalRuleML, conditional deontic norms are represented using specialized rules
called \texttt{PrescriptiveStatement}s of the form \ldots
\begin{verbatim}
<lrml:PrescriptiveStatement>
  <ruleml:Rule closure="universal">
    <ruleml:if>
      ...
    </ruleml:if>
    <ruleml:then>
      ...
    </ruleml:then>
  </ruleml:Rule>
</lrml:PrescriptiveStatement>
\end{verbatim}
where both the \texttt{if}-node (the body) and the \texttt{then}-node (the head) may contain
the LegalRuleML deontic operators \texttt{Obligation}, \texttt{Permission}
and \texttt{Prohibition}, and combinations thereof using the usual connectives.
The semantics of the deontic operators is left underspecified by LegalRuleML, so that any deontic logic may be assumed, e.g., via the \texttt{appliesModality} edge element, to interpret the represented
norms. \texttt{ConstitutiveStatement}s represent so-called \emph{counts-as} norms, and they cannot have
deontic operators in their head. 

Of course, there are many more relevant notions for normative
reasoning (and representable in LegalRuleML) that are not shown here. For a
thorough introduction to LegalRuleML, we refer to the literature~\cite{AthanGPPW15} and the LegalRuleML core specification~\cite{legalruleml}

\subsection{Domain-specific Languages} \label{subsec:dsl}

Domain-specific languages (DSLs) are formal languages (e.g.\ programming or logical languages) that have been designed for use in a particular domain. Their expressivity is deliberately restricted to allow for a higher degree of abstraction, and thus to better leverage specialized domain knowledge of their users.

DSLs can be divided into stand-alone and embedded \cite{mernik2005and}. The former provide their own custom syntax and semantics, thus allowing for a maximal level of customization, but represent a significant implementation effort by requiring the provision of a complete compilation tool chain (parser, type-checker, etc.). The latter consist essentially in a collection of definitions encoded using a more expressive ‘host’ language; this way the existing infrastructure and tools of the host environment can be reused for the DSL. In this case we often speak of an object language (the DSL) that has been embedded into the host language. 

In the context of embedded DSLs, one can further differentiate between two embedding techniques, termed \textit{deep} and \textit{shallow} embeddings \cite{DeepShallow}.
In a deep embedding, the terms of the object language are encoded as inductive data structures in the host language, i.e., as its abstract syntax tree (AST), and term interpretation functions (providing the semantics) can then be defined inductively, e.g.\ for evaluation/execution or optimization. In contrast, terms in shallow embeddings correspond to syntactic abbreviations of the host language, and thus directly encode the intended semantics of object-language expressions. Hence, evaluation in a shallow embedding corresponds to evaluation in its host language, bypassing the need for defining and inductively traversing an AST. In the context of (non-classical) logic a special technique, termed \textit{shallow semantical embeddings} \cite{J41}, has been developed to harness shallow embeddings to encode (quantified) non-classical logics into classical higher-order logic.

Examples of DSLs are query languages for databases such as SQL, hardware design languages such as VHDL and Verilog, scripting languages for game engines and music programming, etc.

\section{Logical Pluralism in Normative Reasoning} \label{sec:logic-pluralism}

\subsection{The Problem of Formalization}
The problem of finding adequate logical formalizations for natural-language discourse is a rather well-known problem in philosophy, where theoretical investigations have shed light on the often underestimated complexity of this endeavor, often putting into question the traditionally strict separation between logical and extra-logical (resp.~categorematic and syncategorematic) expressions, whereby the interpretation of logical connectives can become an additional degree of freedom in the formalization process (cf.~\cite{peregrin2017reflective} and references therein).

In computer science, the idea of mechanistically computing formal representations of natural language in a purely compositional way, made popular through the seminal work of Richard Montague \cite{montague1970universal}, has been pursued with the help of automated reasoners during the last thirty or so years in the area known as \textit{computational semantics} \cite{blackburn2005representation}.\footnote{Much of its development can be followed in the four volumes of the Springer book series \textit{Computing Meaning} from the years 1999, 2001, 2008, and 2013, respectively.} One of the main insights has been that the expressions in natural language are semantically underspecified in the sense that not enough information can be extracted from them to construct the sort of meaning representations Montague was dreaming of, that is, formulas in some formal logical language. Thus, interpreting ordinary sentences can lead to an unfeasible number of different meaning representations \cite{bunt1999computational}.

Among the main determinants behind this underspecification phenomenon, we find ambiguity (syntactic and semantic) and the lack of background knowledge. Among the proposed solutions to tackle the first issue, several kinds of \textit{underspecified semantic representations} have been proposed. However, they have been seen as challenging the application of automated reasoning methods, since disambiguation often results in  different formalizations licensing disparate sets of inferences \cite{10.2307/40180356}. This has been seen commonly as a problem according to the traditional conception that each natural-language statement shall be correlated with one most adequate (`correct') formalization.

On the other hand, the available (formalized) background knowledge is also a degree of freedom determining which inferences are to be drawn from a formalized set of sentences. This knowledge can be of a linguistic nature (e.g.\ lexica) or more domain-specific (e.g.\ ontologies and knowledge bases). In fact, the availability of adequate sources for background knowledge is a well-known bottleneck in the \textit{computational semantics} endeavor. In this respect, RuleML and related knowledge representation and interchange standards, in particular LegalRuleML, play a fundamental role in enabling the interfacing with available normative knowledge sources and ontologies.

\subsection{Formalizing Normative Discourse}

The problem of formalization depicted above applies notably to the logical encoding of normative discourse. This has been experienced with particular intensity in the area of deontic logic, which has witnessed a continuous cat-and-mouse game between sophisticated deontic logical systems designed to avoid some well-known reasoning `paradoxes', and ingenious counterexamples to the claims of the former, which often give rise to new puzzles.
An example of the above is the Chisholm's paradox, as presented in Sect.~\ref{subsec:deontic-logics}, where we could appreciate how the task of adequately formalizing a set of simple natural-language sentences can give rise not only to different logical forms, but also to different ways of interpreting logical connectives, such as conditionals, (deontic) modalities, etc.

In this work, we aim at doing justice to the complex problem of formalizing normative discourse, and thus suggest to employ normative domain-specific languages (DSLs) as an intermediate representation format to encode normative knowledge in a \textit{semantic underspecified} fashion, even reaching to the level of the logical connectives themselves, which thus require further specification for subsequent reasoning tasks. This introduces a component of \textit{logical pluralism} into our approach, since (semantically underspecified) logical operators can (and will) be given concrete interpretations in different non-classical logics, see Sect.~\ref{sec:backends}.
Moreover, we aim at showing not only that such a DSL can (and should) be of a formal logical nature, but also that it can at the same time be fully machine-readable for subsequent consumption by automated reasoning tools. Hence we introduce an illustrative normative DSL \textit{embedded} in a suitable TPTP language below.

\section{Normative Knowledge Representation in the TPTP \label{sec:normativeKR}}

TPTP traditionally focused on classical logic, e.g., standards and benchmark sets for classical propositional and (first- and higher-order) predicate logic formalisms.
Only recently, there have been some ongoing efforts on extending TPTP towards non-classical logics as well.\footnote{See the respective TPTP proposal at
\url{https://tptp.org/NonClassicalLogic/}.} For this purpose, the TFF and THF languages briefly introduced in Sect.~\ref{subsec:tptp}
have been extended with 
expressions
of the form \ldots \\
\mbox{} \quad {\tt \{}{\em connective\_name}{\tt \} @ }{\tt (}{\em arg$_1$}{\tt ,}\ldots{\tt,}{\em arg$_n$}{\tt )}\\
in the first-order TFF language, and \ldots \\
\mbox{} \quad {\tt \{}{\em connective\_name}{\tt \} }{\tt @ }{\em arg$_1$ }{\tt @ }\ldots{\tt @ }{\em arg$_n$}\\
in the higher-order THF language, 
where {\em connective\_name} is either a TPTP-defined name (starting with a {\tt \$} sign)
or a user-defined name (starting with two {\tt \$} signs) for a non-classical operator, and
the {\em arg$_i$} are terms or formulas to which the operator is applied. TPTP-defined connectives
have a fixed meaning and are documented by the TPTP; the interpretation of user-defined connectives is provided by third-party systems, environments, or documentation. 
Non-classical operators may optionally be parameterized with key-value arguments (see below for exemplary use). The so enriched
TPTP languages are denoted NXF (non-classical extended first-order form) and
NHF (non-classical higher-order form), extending TFF and THF respectively.

Non-classical logic languages often come with different logics (e.g., different semantics) associated with them. A prominent example are modal logic languages
in which the properties of the box operator $\Box$ depend on the concrete modal logic
at hand~\cite{BBW2007}. For example, in modal logic \textbf{S5} all instances of $\Box \varphi \rightarrow \varphi$ are tautologies, while this is not the case in modal logic \textbf{K} -- still both logics share the same vocabulary. In order to resolve these
ambiguities and to specify
the exact logic under consideration, non-classical TPTP adds \emph{logic specifications}
to the language. They are annotated formulas of form (here: in NXF) \ldots \\
\mbox{} \quad {\tt tff(}{\em name}{\tt , logic,} {\em logic\_name }{\tt == [}{\em options}{\tt ]} {\tt ).} \\
where {\tt logic} is the TPTP role, {\em logic\_name} is a TPTP-defined or user-defined designator for a logical
language and {\em options} are comma-separated key-value pairs that fix the
specific logic based on that language.
An in-depth overview of the new format is presented in~\cite{SF+22}.
Of course, changing the \textit{logic specification} may change the provability/validity
of the underlying reasoning problem.

The NXF problem representing the formalization (SDL-v3) of 
Chisholm's paradox, as introduced in Sect.~\ref{subsec:deontic-logics}, in
simple modal logic \textbf{D} is as follows (where \verb|{$box}| represents the modal box operator, denoted $O$ in SDL):
\begin{verbatim}
  tff(spec, logic, $modal == [$modalities == $modal_system_D, ...]
  
  tff(norm1, axiom, {$box} @ (help)).
  tff(norm2, axiom, help => {$box} @ (tell)).
  tff(norm3, axiom, ~help => {$box} @ (~tell)).
  tff(fact1, axiom, ~help).
\end{verbatim}
The first line specifies the modal logic to be used (here, modal logic \textbf{D}), while the remaining four lines encode the formulas from 
Sect.~\ref{subsec:deontic-logics}. For illustration purposes, not all of the logic parameters are shown in the \textit{logic specification}.
A list of logics supported by the TPTP so far, their parameters, and their representation is available in the literature~\cite{SF+22}.

Note that both NXF and NHF are quantified languages, allowing for first-order and higher-order quantification, respectively,
within the problem representation. The above example of Chisholm's paradox, however, does not make use of this expressivity.

\subsection{NMF: A Normative DSL in TPTP \label{ssec:nmf}}
The non-classical TPTP formats introduced above allow for encoding non-classical logics for use with 
generic ATP systems. Nevertheless, each problem representation in that format needs to have a fixed underlying logic as
specified by the \textit{logic specification} \cite{SF+22}.
In the present work we want to allow for working with many different normative logics in a uniform way; for this sake we introduce an \emph{embedded DSL} (Sect.~\ref{subsec:dsl})
hosted on top of non-classical TPTP formats, and referred to as \emph{Normative Meta Form} (NMF)
in the remainder.
This way, every file represented in NMF will be syntactically well-formed TPTP, and
hence we can use standard TPTP tools, such as syntax checkers, for processing them. Also,
available software packages for ATP systems, e.g.\ parsers, can be reused.

More specifically, NMF extends NXF from above as follows:
The operator names \verb|$$obligation|, \verb|$$permission|,
\verb|$$prohibition|, and \verb|$$constitutive| are introduced.
They are binary operators, and interpreted as follows \ldots
\begin{itemize}
    \item \verb|{$$obligation} @ (body, head)| encodes that 
    \verb|head| is obligatory given \verb|body|,
    \item \verb|{$$permission} @ (body, head)| encodes that
    \verb|head| is permitted given \verb|body|,
    \item \verb|{$$prohibition} @ (body, head)| encodes that
    \verb|head| is prohibited given \verb|body|,
    \item each of the three deontic operators may optionally be parameterized with the \verb|bearer| option,
    e.g. \verb|{$$obligation(bearer := x)} @ (body, head)|,
    to denote a directed deontic statement towards entity \verb|x|, and
    \item \verb|{$$constitutive} @ (body, head)| encodes a constitutive norm (counts-as norm) that establishes the institutional fact
    that \verb|body| counts as \verb|head|.
\end{itemize}
Since NMF extends NXF, it does not come with a fixed logic  and is thus semantically underspecified. 
We can choose a concrete interpretation of the underspecified deontic
operators using the \textit{logic specification} as follows ...\\
\mbox{} \quad  {\tt tff(}{\em name}{\tt , logic, \$\$normative == [ \$\$logic ==} {\em target\_logic }{\tt ]).}\\
where {\em target\_logic} is some deontic logic identifier. We will describe the target logics currently supported
in Sect.~\ref{sec:backends}. For the time being, it is important to highlight that the description of the encoded norms will remain
the same, regardless of which target logic we choose, and we only need to give a \textit{logic specification} for the desired logic.
Note that some deontic logics, such as SDL, do not come with built-in operators for conditional deontic expressions.
Hence, our normative DSL (NMF) has been designed to abstract away the deontic operators of concrete logics,
and we show in Sect.~\ref{sec:backends} how to translate from NMF to concrete deontic logics.

The running example of Chisholm's paradox can be encoded in NMF in a logically underspecified way
(i.e.\ without a \textit{logic specification}) as follows:
\begin{verbatim}
  tff(norm1, axiom, {$$obligation} @ ($true, help)).
  tff(norm2, axiom, {$$obligation} @ (help, tell)).
  tff(norm3, axiom, {$$obligation} @ (~help, ~tell)).
  tff(fact1, axiom, ~help).
\end{verbatim}

Recall that NMF is defined on top of NXF and, as such, offers first-order quantification, predicate symbols and function symbols.
An even more expressive, higher-order quantified, variant of NMF could be defined analogously on top of NHF (not discussed here).

\subsection{Conversion from LegalRuleML to NMF} \label{subsec:lrml2tptp}
The top-level LegalRuleML statements are translated into NMF as presented in Table~\ref{table:lrml2tptp}.
Note that we are currently addressing only a small fragment of LegalRuleML with this translation; many important metadata present in the
LegalRuleML documentation are not yet considered. In this initial stage we primarily target automation of normative codes as formalized using deontic logics.
In particular, suborder lists are currently not supported, and also strengths/exception specifications of deontic statements are not yet captured. 

\begin{table}[th!]
\centering
\noindent\begin{tabular}{p{.48\textwidth}|p{.48\textwidth}}
 \multicolumn{1}{c|}{\bfseries LegalRuleML}  &  \multicolumn{1}{c}{\bfseries NMF} \\ \hline
 \begin{minipage}{\textwidth}
   \begin{lstlisting}[basicstyle=\scriptsize\ttfamily,mathescape]
<lrml:PrescriptiveStatement key="$\mathit{id}$">
  <ruleml:Rule closure="$\mathit{cl}$">
    <ruleml:if> $\mathit{formula}_1$ </ruleml:if>
    <ruleml:then>
      <lrml:Obligation>
        $\mathit{formula}_2$
      </lrml:Obligation>
    </ruleml:then>
  </ruleml:Rule>
</lrml:PrescriptiveStatement>\end{lstlisting}
\end{minipage} & 
 \begin{minipage}{\textwidth}
   \begin{lstlisting}[basicstyle=\scriptsize\ttfamily,mathescape]
tff($\mathit{id}$, axiom,
    $Q$ [$V_1$, $\ldots$, $V_n$] :
      {$\textdollar\textdollar$obligation} @ (
        $\mathit{tr}(\mathit{formula}_1)$, 
        $\mathit{tr}(\mathit{formula}_2)$ ) ).
\end{lstlisting} 
\end{minipage}
   \\ \hline
 \begin{minipage}{\textwidth}
   \begin{lstlisting}[basicstyle=\scriptsize\ttfamily,mathescape]
<lrml:PrescriptiveStatement key="$\mathit{id}$">
  <ruleml:Rule closure="$\mathit{cl}$">
    <ruleml:if> $\mathit{formula}_1$ </ruleml:if>
    <ruleml:then>
      <lrml:Permission>
        $\mathit{formula}_2$
      </lrml:Permission>
    </ruleml:then>
  </ruleml:Rule>
</lrml:PrescriptiveStatement>\end{lstlisting}
\end{minipage} & 
 \begin{minipage}{\textwidth}
   \begin{lstlisting}[basicstyle=\scriptsize\ttfamily,mathescape]
tff($\mathit{id}$, axiom,
    $Q$ [$V_1$, $\ldots$, $V_n$] :
      {$\textdollar\textdollar$permission} @ (
        $\mathit{tr}(\mathit{formula}_1)$, 
        $\mathit{tr}(\mathit{formula}_2)$ ) ).
\end{lstlisting} 
\end{minipage}
   \\ \hline
 \begin{minipage}{\textwidth}
   \begin{lstlisting}[basicstyle=\scriptsize\ttfamily,mathescape]
<lrml:PrescriptiveStatement key="$\mathit{id}$">
  <ruleml:Rule closure="$\mathit{cl}$">
    <ruleml:if> $\mathit{formula}_1$ </ruleml:if>
    <ruleml:then>
      <lrml:Prohibition>
        $\mathit{formula}_2$
      </lrml:Prohibition>
    </ruleml:then>
  </ruleml:Rule>
</lrml:PrescriptiveStatement>\end{lstlisting}
\end{minipage} & 
 \begin{minipage}{\textwidth}
   \begin{lstlisting}[basicstyle=\scriptsize\ttfamily,mathescape]
tff($\mathit{id}$, axiom,
    $Q$ [$V_1$, $\ldots$, $V_n$] :
      {$\textdollar\textdollar$prohibition} @ (
        $\mathit{tr}(\mathit{formula}_1)$, 
        $\mathit{tr}(\mathit{formula}_2)$ ) ).
\end{lstlisting} 
\end{minipage}
   \\ \hline
   \begin{minipage}{\textwidth}
   \begin{lstlisting}[basicstyle=\scriptsize\ttfamily,mathescape]
<lrml:ConstitutiveStatement key="$\mathit{id}$">
  <ruleml:Rule closure="$\mathit{cl}$">
    <ruleml:if> $\mathit{formula}_1$ </ruleml:if>
    <ruleml:then>
      $\mathit{formula}_2$
    </ruleml:then>
  </ruleml:Rule>
</lrml:ConstitutiveStatement>\end{lstlisting}
\end{minipage}
&
 \begin{minipage}{\textwidth}
 \begin{lstlisting}[basicstyle=\scriptsize\ttfamily,mathescape]
tff($\mathit{id}$, axiom,
    $Q$ [$V_1$, $\ldots$, $V_n$] :
      {$\textdollar\textdollar$constitutive} @ (
        $\mathit{tr}(\mathit{formula}_1)$, 
        $\mathit{tr}(\mathit{formula}_2)$ ) ).
\end{lstlisting} 
\end{minipage}
\\ \hline
\begin{minipage}{\textwidth}
\begin{lstlisting}[basicstyle=\scriptsize\ttfamily,mathescape]
<lrml:FactualStatement key="$\mathit{id}$">
  $\mathit{formula}_1$
</lrml:FactualStatement>\end{lstlisting}
\end{minipage}
&
\begin{minipage}{\textwidth}
\begin{lstlisting}[basicstyle=\scriptsize\ttfamily,mathescape]
tff($\mathit{id}$, axiom, $\mathit{tr}(\mathit{formula}_1)$ ).
\end{lstlisting}
\end{minipage}
\\
\end{tabular}
\caption{Translation scheme from a fragment of LegalRuleML to NMF.
In each case except the last one the quantification closure of the formula is explicitly added to the TPTP translation; so that
$\{V_1, \ldots, V_n\} = \mathrm{fv}(\mathit{formula}_1) \cup \mathrm{fv}(\mathit{formula}_2)$ and $Q = \mathtt{!}$ if $cl = \mathtt{universal}$
and $Q = \mathtt{?}$ if $cl = \mathtt{existential}$. The explicit quantification is omitted if $n = 0$. $\mathit{tr}(.)$ is an adequate mapping
from RuleML formulas to TPTP formulas. \label{table:lrml2tptp}
}
\end{table}

The translation process recursively translates the prescriptive statements, constitutive statements and factual statements of LegalRuleML 
into formulas in NMF. For identification purposes, key references from LegalRuleML are kept as formula names in the TPTP representation,
and additional (legal) references and associations, expressed via \verb|<lrml:LegalReferences>| or \verb|<lrml:References>| blocks, and assigned
by \verb|<lrml:Associations>| blocks, respectively, are kept during the translation as TPTP annotations (not shown in Table~\ref{table:lrml2tptp}).
If a deontic operator in LegalRuleML comes with a \verb|<lrml:Bearer>| node, this is mirrored in NMF as sketched in Sect.~\ref{ssec:nmf}.

The translation from LegalRuleML to the proposed logic-pluralistic TPTP-based DSL is prototypically implemented as part of the
\emph{tptp-utils} tool, available at GitHub\footnote{See \url{https://github.com/leoprover/tptp-utils} and its README there.}.
\emph{tptp-utils} will produce a NMF file according to the above translation scheme but without
a \textit{logic specification}. The latter can be added by the user in order to assume concrete interpretations of the normative
statements, see Sect.~\ref{sec:backends}. A \textit{logic specification} could also be 
created automatically from the LegalRuleML document, deriving
from respective \verb|appliesModality| edges; this is an interesting venue for further work.

\section{TPTP-based Normative Reasoning Backends \label{sec:backends}}

\begin{figure}[tb]
  \centering
  \includegraphics[width=.9\textwidth]{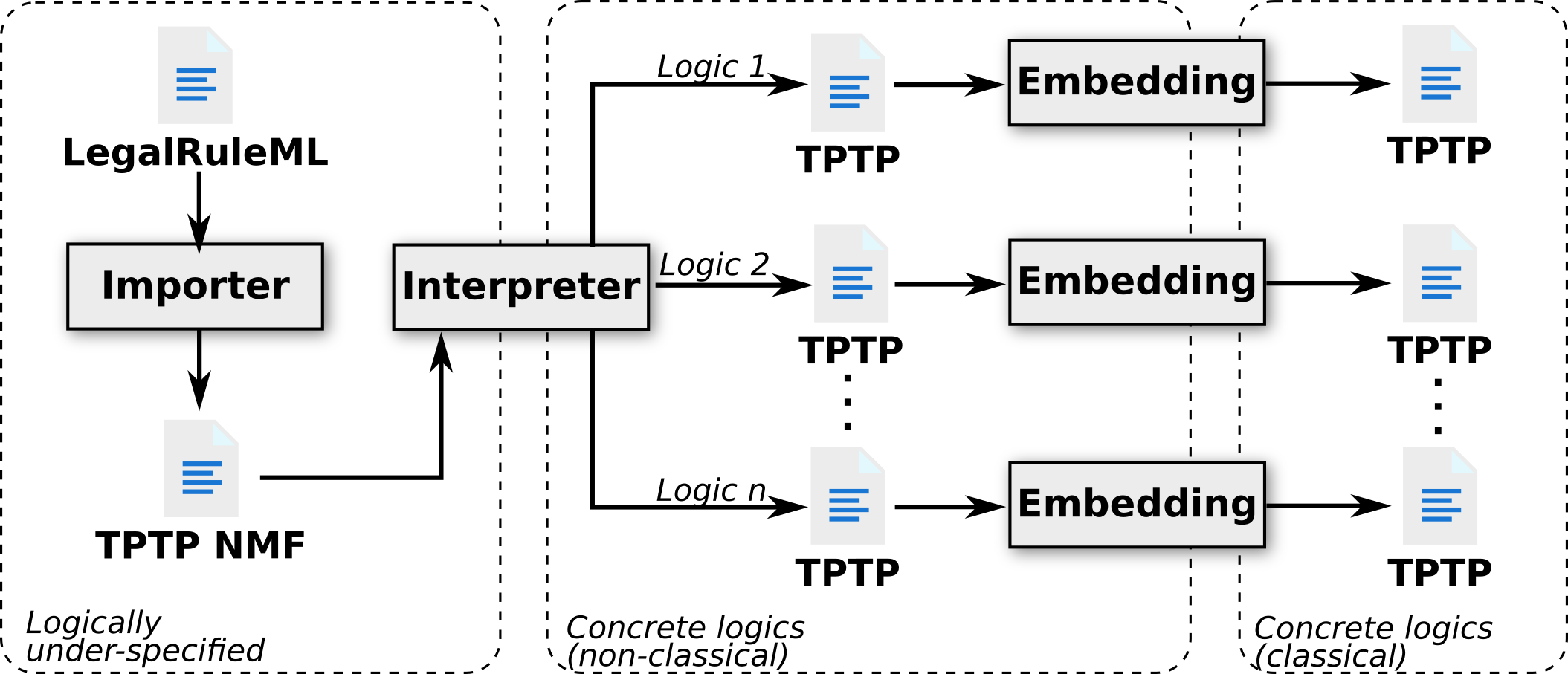}
  \caption{Visualization of the transformation and automation process. One LegalRuleML file
  can be translated into multiple different TPTP reasoning problems which, in turn, can be
  reduced to classical reasoning problems in HOL for automation, if no special-purpose prover of
  the desired target logic is available.}
  \label{fig:overview}
\end{figure}

The translation of LegalRuleML statements into a representation in the TPTP-based DSL introduced above does not yet allow
the utilization of automated reasoning tools for automated normative reasoning. It does give, though, an abstract
representation of the encoded information in a format that we can use to provide means for automation via the general
TPTP automated reasoning infrastructure. To this end, two steps are necessary: (i) The transformation of the encoded
norms into a concrete (deontic) logical formalism, and (ii) the provision of ATP systems that can reason within the
respective logics.

The overall approach for logic-pluralistic automated reasoning presented in this paper
is visualized in Fig.~\ref{fig:overview}. The translation of LegalRuleML into NMF (Sect.~\ref{subsec:lrml2tptp}) connects
normative knowledge representation to the TPTP infrastructure, where both LegalRuleML and NMF are logically underspecified.
Subsequently, the NMF representation is translated to multiple concrete logic problems formulated in (standard) non-classical TPTP.
These problems are not logically underspecified anymore, as they have been encoded into specific deontic logics.
Then, the resulting non-classical problems are automated using the \textit{shallow semantical embeddings} approach \cite{J41}, in which
the problems are encoded into classical higher-order logic (HOL)~\cite{DBLP:conf/ecai/FuenmayorB20,DBLP:journals/flap/BenzmullerFMP19,DBLP:journals/flap/BenzmullerFP19,DBLP:conf/deon/BenzmullerFP18,DBLP:conf/ruleml/GleissnerS18}. This way, general purpose HOL ATP solvers can be employed
for normative reasoning. Of course, also specialized ATP systems for the respective deontic logic could be employed.
However, for many quantified non-classical logics there are no ATP systems available.

The NMF representation is interpreted with respect to a concrete
logic by adding a \textit{logic specification} to it. It is of form \ldots \\
\mbox{}\quad {\tt tff(}{\em name}{\tt , logic, \$\$normative == [\$\$logic == }{\em target\_logic}{\tt ]).}\\
where {\em target\_logic} is one of ...
\begin{itemize}
    \item {\tt \$\$sdl}, representing SDL as introduced above,
    \item {\tt \$\$aqvistE}, representing {\AA}qvist dyadic deontic logic \textbf{E}~\cite{aaqvist1984deontic}, and 
    \item {\tt \$\$carmoJones}, representing the dyadic deontic logic of Carmo and Jones~\cite{carmo2002deontic}.
\end{itemize}
Of course, this list can be extended with many more concrete logics for deontic reasoning. For the proof-of-concept
presented in this paper, we restrict ourselves to these logics for the time being.
An NMF problem with a \textit{logic specification} can then be translated to a non-classical TPTP representation
of the respectively chosen logic.
The translation schemes for translating NMF into SDL and into DDL are presented in Tables~\ref{table:NMF2SDL} and~\ref{table:NMF2DDL}.

\begin{table}[t]
    \centering
    \begin{tabular}{l|l}
        \multicolumn{1}{c|}{\bfseries NMF}  &  \multicolumn{1}{c}{\bfseries SDL} \\ \hline    
        \verb|{$$obligation} @ (body, head)|  & \verb|body => {$box} @ (head)|  \\
        \verb|{$$permission} @ (body, head)|  & \verb|body => {$dia} @ (head)|  \\
        \verb|{$$prohibition} @ (body, head)|  & \verb|body => {$box} @ (~head)|  \\
        \verb|{$$constitutive} @ (body, head)|  & \verb|body => head|  \\
    \end{tabular}
    \smallskip
    
    \caption{Translation of deontic operators from NMF to SDL based on the SDL-v3 scheme from Sect.~\ref{subsec:deontic-logics} (narrow scope). Directed deontic operators
    are modeled, in each case, via {\tt \$box(\#x)} resp. {\tt \$dia(\#x)} where {\tt x} is the bearer of the modality.}
    \label{table:NMF2SDL}
\end{table}

\begin{table}[t]
    \centering
    \begin{tabular}{l|l}
        \multicolumn{1}{c|}{\bfseries NMF}  &  \multicolumn{1}{c}{\bfseries DDL} \\ \hline    
        \verb|{$$obligation} @ (body, head)|  & \verb|{$$obl} @ (head, body)|  \\
        \verb|{$$permission} @ (body, head)|  & \verb|~{$$obl} @ (~head, body)|  \\
        \verb|{$$prohibition} @ (body, head)|  & \verb|{$$obl} @ (~head, body)|  \\
        \verb|{$$constitutive} @ (body, head)|  & \verb|body => head|  \\
    \end{tabular}
    \smallskip
    
    \caption{Translation of deontic operators from NMF to DDL. Directed deontic operators
    are not yet supported.}
    \label{table:NMF2DDL}
\end{table}

In SDL the obligation operator is expressed using the modal logic $\Box$ operator; and the logic is specified to be modal logic \textbf{D}
(as usual for SDL). Since SDL does not have any dyadic deontic operators, conditional norms are expressed via a material implication.
DDL does provide a dyadic deontic operator that captures conditional norms, so the mapping is more natural here.
Note that, for simplicity, the translation scheme currently follows the interpretation variant SDL-v3 (see Sect.~\ref{subsec:deontic-logics}) using a narrow-scope
translation. It is planned to add further parameters to the translation so that the translation scheme can be chosen
individually for each norm.

For the running example of Chisholm's paradox, as formalized in NMF in Sect.~\ref{ssec:nmf},
the concrete output for SDL as reasoning target is as follows:

\begin{verbatim}
  tff(target, logic, $modal == [$quantification == $constant,
                                $constants == $rigid,
                                $modalities == $modal_system_D]).

  tff(norm1-sdl, axiom, {$box} @ (help)).
  tff(norm2-sdl, axiom, help => {$box} @ (tell)).
  tff(norm3-sdl, axiom, ~help => {$box} @ (~tell)).
  tff(fact1-sdl, axiom, ~help).
\end{verbatim}
In {\AA}qvist system \textbf{E} the resulting representation is
(note the different order of parameters in the dyadic deontic operator) \ldots
\begin{verbatim}
  tff(target, logic, $$ddl == [$$system == $$aqvistE]).

  tff(norm1-ddl, axiom, {$$obl} @ (help,$true)).
  tff(norm2-ddl, axiom, {$$obl} @ (tell,help)).
  tff(norm3-ddl, axiom, {$$obl} @ (~tell,~help)).
  tff(fact1-ddl, axiom, ~help).
\end{verbatim}
For the DDL of Carmo and Jones, the output is identical except that the \textit{logic specification} 
gives \verb|$$carmoJones| instead of \verb|$$aqvistE|. For details on the non-classical logics supported
by the TPTP and the deontic logics used above, we refer to the literature~\cite{Ste22,SF+22}.
Note that in all three cases, the problems have a fixed semantics and can thus be processed by ATP systems.
The presented translation process from NMF to the deontic logics is implemented in the \textbf{LET} tool for logic embeddings~\cite{Ste22},
also available at GitHub\footnote{See \url{https://github.com/leoprover/logic-embedding} and its README there.}.

In a second step, the NXF problems are embedded into classical HOL problems, represented in the THF TPTP-format.
This is also done via the \textbf{LET} tool. We refer to the literature for details on this process~\cite{Ste22}.
The automation of normative reasoning via shallow embedding into HOL, as illustrated by Benzm\"uller et al.\
via their LogiKEy methodology~\cite{BenzmullerPT20}, has been successful for a broad range of applications~\cite{J41}.

In order to provide a seamless automation process, the \textbf{LET} tool has been included included into the
higher-order ATP system Leo-III~\cite{DBLP:journals/jar/SteenB21}, so that the above problem statements in SDL and DDL can be given 
to Leo-III without the need for any external pre-processing via \textbf{LET} by the user.
Unsurprisingly, Leo-III can automatically establish the unsatisfiability of the four SDL formulas \verb|norm1-sdl|,
\verb|norm2-sdl|, \verb|norm3-sdl| and \verb|fact1-sdl|, thus proving their joint inconsistency; by contrast consistent conclusions can be drawn from the DDL representation.

\section{Conclusion \label{sec:conclusion}}

In this paper we presented a flexible approach for using general-purpose (classical) ATP systems for normative reasoning.
This is motivated, on the one hand, by the widespread availability of mature and practically effective ATP systems, and, on the other-hand,
by practical challenges for providing ATP systems for the many different deontic logics employed in normative reasoning.
Hence, we aim at bridging between the ATP systems community (users of the TPTP problem representation languages)
and the normative knowledge representation and reasoning community (users of the LegalRuleML standard).

Our proposed approach consists in first translating a subset of LegalRuleML to a specifically crafted domain-specific language, denoted NMF,
based on the TPTP standard for ATP systems. NMF is semantically underspecified and acts as an intermediate layer
between natural-language representations and representations in specific logical formalisms. 
NMF is subsequently
translated into different reasoning problems in concrete logics, represented in the recent non-classical TPTP standard. 
Finally, automation for non-classical TPTP is provided by \textit{shallow semantical embeddings} into classical higher-order logic for
which many different ATP systems exist.
While from a purely conceptual knowledge representation perspective the intermediate NMF language might not be strictly necessary
(i.e., LegalRuleML could be translated directly into concrete TPTP problem), we argue that the
usage of a semantically underspecified TPTP-based representation language comes with several pragmatic advantages with respect
to practical automation. The TPTP languages are the standard formats for automated reasoning and experimentation using ATP systems,
hence lowering the engineering-related barriers of providing automation for different deontic logics, and at the same time providing
an abstract language for experimentation with different logics and ATP systems. In particular, the standard TPTP problem library~\cite{Sut17}
for ATP system evaluation collects abstract problems (so-called \emph{generators}) from which concrete reasoning tasks can be
generated. The NMF layer thus connects to these efforts by providing means for the logic-pluralistic representation of domain-specific
reasoning benchmarks.

The different steps in this process have been implemented as open-source tools available at GitHub.
By doing so, we provided a flexible reasoning infrastructure for logic-pluralistic normative reasoning
that is in line with the LogiKEy methodology~\cite{BenzmullerPT20} for designing normative theories.
In contrast to LogiKEy, our focus is on flexible reasoning via ATP systems instead of enabling the interactive use
of proof assistants. Two examples, one of them being the
discussed Chisholm's paradox, are available as supplemental dataset via Zenodo~\cite{steen_alexander_2022_6702576}. The dataset contains the initial LegalRuleML documents, their NMF representations and all translations to the three concrete logics.

\paragraph{Further Work.}
In this paper, we focused on three deontic logics as reasoning backends for NMF. We plan to extend the portfolio of supported
deontic logics towards further relevant ones, including Input/Output logic~\cite{makinson2000input}. Also, a more fine-grained control
of the translation schemes from NMF to the respective logics can be achieved when allowing to choose between different alternatives
for each norm. 
It is planned to extend the translation tool to produce LegalRuleML output from deontic logic reasoning problems formulated in TPTP.

The DSL presented in this work is still prototypical. It does not yet capture many important aspects that are encoded in LegalRuleML documents.
In particular, it is possible to extend our approach to a layered hierarchy of DSLs aiming for knowledge representation
at different levels of abstraction (or domain-specificity), together with translation mechanisms for successively specifying
their intended semantics. Furthermore, RuleML does allow to specify so-called \emph{semantic profiles}
\footnote{See \url{https://wiki.ruleml.org/index.php/Semantic_Attributes_and_Semantic_Profiles}.}
to encode the intended logic (or: semantics) under which the input should be interpreted.
It seems a fruitful venue to closer connect these profiles to TPTP logic specifications in order to allow for a more principled
approach to adjust the target logic in automation. 
Also, there are ongoing efforts to extend the TPTP to allow for structured usage
of name space hierarchies, e.g. for ontology inclusion.\footnote{See \url{https://tptp.org/TPTP/Proposals/IncludeNameSpaceHierarchy.html}.}.
It is planned to connect to these efforts to align our approach with well-known ontology standards.

\subsubsection{Acknowledgements} We a grateful for the constructive feedback from the anonymous reviewers which
greatly improved this work.

\bibliographystyle{splncs04}
\bibliography{main-arxiv}

\end{document}